%% file: neurips_2025.tex
\pdfoutput=1
\documentclass{article}

\usepackage[preprint]{neurips_2025}

\usepackage[utf8]{inputenc} %
\usepackage[T1]{fontenc}    %
\usepackage{hyperref}       %
\usepackage{url}            %
\usepackage{booktabs}       %
\usepackage{amsfonts}       %
\usepackage{nicefrac}       %
\usepackage{microtype}      %
\usepackage[table,xcdraw]{xcolor} %

\input{pkgs}

\input{cmds}

\title{\scalebox{0.78}{AutoSpec: Automated Generation of Neural Network Specifications}}

\author{
    \textbf{Shuowei Jin\textsuperscript{1}, 
    Taobo Liao\textsuperscript{4},
    Anuj Kalia\textsuperscript{2},
    Xenofon Foukas\textsuperscript{2},}\\
    \textbf{Huan Zhang\textsuperscript{4},
    Cheng Tan\textsuperscript{3},
    Z. Morley Mao\textsuperscript{1},
    Francis Y. Yan\textsuperscript{4}}
    \vspace{6pt} \\
    \textsuperscript{1}University of Michigan \quad
    \textsuperscript{2}Microsoft Research \\
    \textsuperscript{3}Northeastern University \quad
    \textsuperscript{4}University of Illinois Urbana-Champaign
}

\begin{document}

\maketitle

\input{sections/00-abstract}
\input{sections/01-intro}

\input{sections/02-background2}

\input{sections/03-system-design}

\input{sections/05-evaluation}

\input{sections/07-conclusion}
\clearpage

\bibliographystyle{unsrtnat} 
\bibliography{reference}
\clearpage

\appendix

\input{sections/08-appendix}

\clearpage

\newpage

\end{document}

%% file: pkgs.tex
\usepackage[english]{babel}
\usepackage{blindtext}
\usepackage{xspace}
\usepackage{amsmath}
\usepackage{amsthm}
\usepackage{amssymb} 
\usepackage[normalem]{ulem}
\usepackage{caption}
\usepackage{graphicx}
\usepackage{booktabs}
\usepackage{arydshln}
\usepackage{xurl}
\usepackage{wrapfig}
\usepackage[toc,title]{appendix}
\usepackage{afterpage}
\usepackage{tikz}
\usepackage{enumitem}
\usepackage{stfloats}
\usepackage[ruled,linesnumbered]{algorithm2e}

\SetCommentSty{mycommfont}
\usepackage{comment}
\usepackage{algorithmicx}
\usepackage{algpseudocode}
\usepackage{multirow}

\usepackage{xspace}
\usepackage{mfirstuc} %
\usepackage{longtable}

%% file: cmds.tex
\newenvironment{myitemize2}%
  {\begin{list}{\labelitemi}{\itemsep2pt \topsep2pt \parsep0.00in
  \partopsep=0pt \leftmargin1.2em}}%
  {\end{list}}

\let\latexusecounter=\usecounter
\def\compactsortof{\itemsep=2pt \topsep=2pt \parsep=0.00in \partopsep=0pt
\leftmargin=1.2em}
\newenvironment{myenumerate2}
  {\def\usecounter{\compactsortof\latexusecounter}
   \begin{enumerate}}
  {\end{enumerate}\let\usecounter=\latexusecounter}

\newcommand{\heading}[1]{
  \vspace{0ex}
  \noindent
  \textbf{#1}}

\newcommand{\ttt}[1]{\texttt{#1}}

\newcommand{\sys}{\textsl{AutoSpec}\xspace}
\newcommand{\dt}{\textsl{AutoSpec}\xspace}
\newcommand{\kmeans}{Clustering\xspace}
\newcommand{\grid}{Grid\xspace}
\newcommand{\NaN}{---\xspace}

\newcommand{\mysection}[1]{\vspace{-.01in}\section{#1}\vspace{-.02in}}
\newcommand{\mysubsection}[1]{\vspace{-.06in}\subsection{#1}\vspace{-.03in}}

\newcommand{\spec}{specification\xspace}
\newcommand{\specs}{specifications\xspace}

\newcommand{\CF}[1]{\xmakefirstuc{#1}}

\newtheorem{theorem}{Theorem} 
\newtheorem{proposition}{Proposition}

\def\Snospace~{\S{}}

%% file: sections/00-abstract.tex
\begin{abstract}
The increasing adoption of neural networks in learning-augmented systems highlights the growing need for model safety and robustness, especially in safety-critical domains. While recent advances in neural network verification offer formal guarantees on worst-case behavior, existing approaches require users to manually define model specifications---an error-prone, incomplete, and time-consuming process. In this paper, we present \sys, the first comprehensive framework for automatically generating and evaluating neural network specifications for learning-augmented systems.
\sys introduces a tree-based algorithm that adaptively partitions the input space to generate specification sets aligned with model behavior, as well as a statistical certification framework that provides rigorous accuracy guarantees for each specification. We also propose a principled evaluation framework that defines interpretable metrics for specification accuracy and coverage, establishing a benchmark for future research. Experiments across four diverse applications show that \sys outperforms both manually defined specifications and existing baseline algorithms, improving the F1 score by up to 53\% over human-defined specifications and 73\% over the strongest baseline.
\end{abstract}

%% file: sections/01-intro.tex
\mysection{Introduction}
\label{s:intro}

\CF{Learning-augmented systems}, which integrate neural networks into traditional computer systems, have shown remarkable improvements across various real-world applications, such as database indexing~\cite{kraska2018case}, resource management~\cite{decima}, video streaming~\cite{yan2020learning}, and intrusion detection~\cite{liu2019machine}. Despite these successes, these systems face a significant challenge: neural networks may achieve strong \emph{average} performance but often exhibit unpredictable behavior in \emph{worst-case} scenarios, potentially leading to catastrophic failures. For example, a learned congestion-control algorithm may inappropriately lower packet sending rates even under favorable network conditions~\cite{eliyahu2021verifying}.

Providing formal guarantees for the worst-case performance of learning-augmented systems has recently gained significant attention~\cite{eliyahu2021verifying,wei2023building,wu2022scalable}. Current methods, however, rely heavily on human experts to define the properties to verify, known as \emph{neural network \specs}. Each \spec specifies the expected model outputs for particular input spaces (\autoref{ss:spec-lang}). Traditionally, domain experts manually create these \specs based on intuition and domain knowledge. For instance, in adaptive video streaming, where neural networks select bitrates based on network conditions, \citet{eliyahu2021verifying} define a \spec as: ``[if video] chunks are downloaded faster than playback time, the DNN should eventually avoid selecting the lowest resolution.'' Similar handcrafted \specs have been developed for database indexing~\cite{tan2021building}, memory allocators~\cite{wei2023building}, and job schedulers~\cite{wu2022scalable}.

However, manual \spec creation is inherently problematic due to several reasons. Manually designed \specs typically lack comprehensiveness, addressing only limited aspects of the input domain, leaving large parts of the input space unverified. Handcrafted \specs can also be ambiguous, excessively broad, or incorrect in edge cases, reducing their practical utility. Additionally, manual specification generation demands substantial human expertise and effort, limiting its scalability. These challenges motivate the exploration of automated specification generation, prompting our central research question: \textit{Can we automatically generate \specs directly from data?}

Automatically generating effective specifications from data introduces several unique challenges. There is a critical trade-off between \emph{comprehensiveness} and \emph{accuracy}: broadly-defined specifications risk admitting incorrect behaviors, while overly precise ones might fail to generalize. Achieving an optimal balance through naive data partitioning methods is non-trivial. It is also challenging to produce specifications that provide \emph{formal guarantees} for unseen data without incurring significant computational or data costs. A rigorous statistical approach is needed to convert empirical patterns into reliable guarantees. Furthermore, evaluating generated specifications lacks standardized, principled criteria. Common metrics like coverage and empirical accuracy might inadvertently reward trivial or impractical solutions. Therefore, an effective evaluation framework requires carefully designed metrics that penalize useless specifications.

\heading{Contributions.} In this paper, we introduce \sys, the first comprehensive framework for automated generation and evaluation of neural network specifications.

First, we introduce a tree-based algorithm as the core of our specification generation framework, which leverages decision trees intrinsic characteristics to adaptively partition the input space, capturing both feature and label structures from data. This enables \sys to automatically generate specification sets that are both comprehensive and precisely aligned with models' expected behavior.

Second, we develop a statistical certification framework providing rigorous accuracy guarantees for each specification by utilizing concentration inequalities. This certification ensures each specification robustly represents the underlying data distribution, ensuring reliability and trustworthiness without additional computational or data overhead.

Third, we establish a principled evaluation framework for neural network specifications, defining clear, interpretable metrics such as precision, recall, F1 score, and a certified pass rate. This systematic approach provides a rigorous basis for assessing and comparing specification generation and verification methodologies.

To demonstrate the effectiveness of \sys, we conduct experiments across four diverse applications and perform a case study on anomaly detection in neural networks. Our results show that \sys achieves an F1 score $53\%$ higher than human-defined specifications, and $68\%$ higher than the next best baseline algorithm on average.

%% file: sections/02-background2.tex
\mysection{Related Work}

\heading{Neural Network Verification Frameworks.} Neural network verification is essential for ensuring reliability and trustworthiness in learning-augmented systems. The primary goal of verification is to provide provable guarantees that neural networks satisfy desired \specs and behave as intended. Researchers have developed various verification approaches, including exhaustive search techniques~\cite{bak2022neural}, specialized SMT solvers~\cite{katz2019marabou}, and reachability analysis methods~\cite{wang2021beta}. While these verification frameworks have demonstrated effectiveness, they predominantly rely on manually crafted specifications, which is a labor-intensive and potentially error-prone process. This limitation highlights the critical need for automated specification generation techniques.

\heading{Neural Network Specification Mining.} There are some early explorations to generate specifications for Neural Networks. Geng et al.~\cite{geng2023towards} introduced an approach that mines neural activation patterns (NAPs) as candidate specifications. However, their method is constrained by architectural dependencies and faces scalability challenges with larger models. SpecTra~\cite{chaudhary2024specification} takes a different approach by deriving specifications from existing human-designed algorithms, though this limits its applicability to novel scenarios where such algorithms may not exist. In the broader context, \emph{\spec mining}~\cite{ammons2002mining,lemieux2015general, le2018deep} has been extensively studied in program verification, focusing on automatically formulating specifications for traditional software systems rather than neural networks.

To our knowledge, \sys is the first and only system that provides a comprehensive end-to-end solution for neural network specification generation. It automatically generates specifications directly from data, derives strong theoretical analysis to provide formal bounds, and introduces comprehensive evaluation framework to assess the quality and effectiveness of the generated specifications.

\mysection{Background}

\subsection{Neural Network Specifications}
\label{ss:spec-lang}

Let $\mathcal{N}\!:\mathbb{R}^{n}\!\to\!\mathbb{R}^{m}$ denote a trained
network.  A \emph{specification} is a pair
\(
\phi = (\phi_{X},\phi_{Y})
\)
that constrains inputs and the corresponding outputs:
\begin{equation}
\forall\,x\in\mathbb{R}^{n}:
\quad
\phi_{X}(x)\;\Longrightarrow\;
\phi_{Y}\bigl(\mathcal{N}(x)\bigr).
\label{eq:spec}
\end{equation}
Following the VNN‑LIB benchmark~\cite{demarchi2023supporting}, we focus on generating specifications as axis‑aligned hyper‑rectangles for the input predicate and
simple intervals for the output predicate:
\[
\phi_{X} \;=\; \prod_{p=1}^{n}[l_{p},u_{p}],
\qquad
\phi_{Y} \;=\; [\,y_{\min},\,y_{\max}\,].
\]
This choice balances two goals:  
(i) \textbf{expressiveness}—it can encode many practical safety
properties such as $\ell_\infty$ robustness or bounded control signals;  
(ii) \textbf{efficiency}—state‑of‑the‑art verifiers verify this rectangular form efficiently.  We leave the extending \sys to other types of polytopes (zonotopes, polyhedra, etc.) as future work.
Throughout the paper we use $\Phi=\{\phi_{1},\dots,\phi_{K}\}$ to denote
a \emph{set} of such specifications generated from data
(\autoref{s:specgen}).

\subsection{Hoeffding's Inequality}
\label{ss:hoeffding}

Hoeffding's inequality is a key concentration bound in statistical learning theory, applicable to sums of independent, bounded random variables. It quantifies the probability that an empirical mean deviates from its true expectation by a certain amount.

\begin{theorem}[Hoeffding, 1963]
    \label{thm:hoeffding}
    Let $Z_{1},\dots,Z_{n}$ be independent random variables such that $0\le Z_{i}\le 1$ for all $i$, and let $\mu = \mathbb{E}Z_{i}$ be their common mean. Let $\hat{\mu} = \frac{1}{n} \sum_{i=1}^{n} Z_i$ denote the empirical mean. Then, for every $\varepsilon > 0$:
    \begin{align*}
    &\Pr\bigl[\,|\hat{\mu} - \mu| \geq \varepsilon\,\bigr] \leq 2\exp\bigl(-2n\varepsilon^2\bigr)  \qquad\quad \textnormal{(two-sided)}, \\
    &\Pr\bigl[\,\hat{\mu} - \mu \geq \varepsilon\,\bigr] \leq \exp\bigl(-2n\varepsilon^2\bigr)  \qquad\quad\;\;\;\;\, \textnormal{(one-sided)}.
    \end{align*}
    \end{theorem}

This inequality is foundational for deriving \emph{probably‑approximately‑correct} (PAC) learning guarantees. By interpreting $Z_i$ variables as error indicators for a learning model, Hoeffding's inequality provides a crucial link between the model's empirically observed error (on training data) and its true generalization error (on unseen data). Based on it, we will prove generated specification error bound as discussed in \autoref{s:pac-certify}.

%% file: sections/03-system-design.tex
\input{sections/03-2-specification-generation}

\input{sections/03-3-specification-bound}

\input{sections/03-1-evaluation-metrics}

%% file: sections/03-2-specification-generation.tex
\mysection{Design of \sys}

\input{sections/fig-overview}

\sys\ consists of three main components, as illustrated in \autoref{fig:overview}: 
(1) the \emph{Specification Generation Framework}, which uses the labeled generation dataset to produce a set of candidate specifications; 
(2) the \emph{Specification Certification Framework}, which certifies a formal lower bound on each specification’s accuracy based on empirical observations; and 
(3) the \emph{Specification Evaluation Framework}, which defines a set of interpretable metrics to assess the quality and reliability of the generated specifications. 
These certified and evaluated specifications can then be directly used in downstream verification or monitoring tasks.

\mysubsection{Specification Generation Framework}
\label{s:specgen}
The goal of the generation module is to transform a labeled dataset
\(
\mathcal{D}_{\mathrm{gen}} = \{(x_i, y_i)\}_{i=1}^{N},
\;x_i \in \mathbb{R}^{n},
\;y_i \text{ class or scalar}
\)
into a set of specifications
\(
\Phi = \{\phi_j = (\phi_{X_j}, \phi_{Y_j})\}_{j=1}^{K}
\),
where each $\phi_{X_j} \subseteq \mathbb{R}^n$ is an axis-aligned hyperrectangle in the input space and $\phi_{Y_j} \subseteq \mathbb{R}$ is a corresponding output constraint summarizing the labels in that region, as formalized in \autoref{ss:spec-lang}. 
Additionally, we aim to ensure the usefulness of specifications by optimizing three key properties: (1) accuracy—how well specifications align with ground-truth data, (2) coverage—the extent to which the specification set spans the input space for comprehensive model verification, and (3) constraint tightness—avoiding overly permissive output ranges that would render specifications meaningless. These quality metrics are formally defined and utilized in our evaluation framework in \autoref{s:speceval}.

Our approach to specification generation is guided by a fundamental insight:

\fbox{
    \parbox{0.96\linewidth}{
        \textbf{Insight:} A specification set is an \emph{interpretable, discrete description} of the joint distribution $P(X,Y)$.  
        Each specification carves out a region of the input space and specify the expected output behaviour observed there.  
        Generating these specifications from a finite dataset is analogous to training a model that generalizes from samples to the true distribution.
    }
}

With this understanding, we design three algorithms for generating specifications: an unsupervised geometry-only baseline (grid-based), an unsupervised clustering-based baseline, and our main approach, a supervised decision-tree-based method (\sys). All three methods follow a common workflow: they first partition the input space into hyperrectangles using different strategies, then apply a shared primitive, \textsc{ExtractSpec} (detailed in Algorithm~\ref{algo:spec-extract}), to derive output constraints for each region based on the enclosed data points. This process transforms raw data into formal specifications that capture the underlying distribution patterns.

\heading{Grid-based Generation.}
This baseline algorithm partitions each input dimension into equal-width bins, creating a uniform grid over the input space. For each non-empty cell, we apply \textsc{ExtractSpec}. The detailed algorithm is presented in \autoref{ss:grid}. While simple, this method has two key limitations: (1) it cannot adapt to the underlying data distribution, potentially grouping diverse data points with different labels into a single cell, which leads to imprecise or overly permissive output constraints; and (2) it suffers from the ``curse of dimensionality'' with exponential runtime complexity of $O(\beta^n)$ where $\beta$ is the number of bins per dimension and $n$ is the dimensionality of the input space, making it impractical for high-dimensional data.

\heading{Clustering-based Generation.}
To better adapt to data distribution, we employ clustering algorithms (e.g., $k$-means) on the input points, then form one bounding hyperrectangle per cluster. Each rectangle is processed with \textsc{ExtractSpec} to derive output constraints. While this method reduces redundancy and scales better with dimensionality, it suffers from a fundamental limitation: its unsupervised nature ignores label information during clustering. Consequently, clusters may contain points with diverse labels, leading to overly permissive output constraints that lack specificity. The detailed algorithm is presented in \autoref{ss:cluster}.

\heading{Decision Tree-based Generation (\sys Core Algorithm).}
To overcome the limitations of previous algorithms, especially their insufficient utilization of label information and data distribution, we introduce \sys, our decision-tree based specification generation algorithm that strikes a balance between accuracy and comprehensiveness.

Our design insight is that decision trees are particularly well-suited for specification generation because they naturally partition the input space while considering label distributions. In training a decision tree, we recursively split the input space based on features that best separate the classes (using Gini impurity for classification or variance reduction for regression). This approach creates regions where data points have similar labels, which aligns perfectly with our goal of generating precise specifications with tight output constraints.

The workflow of \sys is as follows: Given a labeled dataset $\mathcal{D}_{\mathrm{gen}}$, we first train a CART decision tree on both input features and corresponding labels. Each path from the root to a leaf node encodes a conjunction of feature predicates (e.g., $0.2 \leq x_1 \leq 0.5 \wedge 0.3 < x_2 \leq 0.7$), naturally forming a hyperrectangle $R_L$ in the input space. For each leaf node's hyperrectangle, we apply \textsc{ExtractSpec} to derive appropriate output constraints based on the enclosed data points, resulting in a formal specification. The complete algorithm is detailed in \autoref{algo:autospec}.

The decision-tree approach offers several significant advantages: (1) it is inherently label-aware, with every split optimized to maximize output consistency within partitions; (2) it adaptively refines the input space, creating fine-grained partitions in regions with complex decision boundaries while maintaining broader partitions where behavior is uniform; and (3) it ensures comprehensive coverage of the data distribution support. Furthermore, \sys achieves high computational efficiency with an average time complexity of $O(Nd\log N)$ for $N$ training examples and $d$ features, making it practical even for high-dimensional datasets.

\begin{figure}[t]
  \centering
  \includegraphics[width=\linewidth]{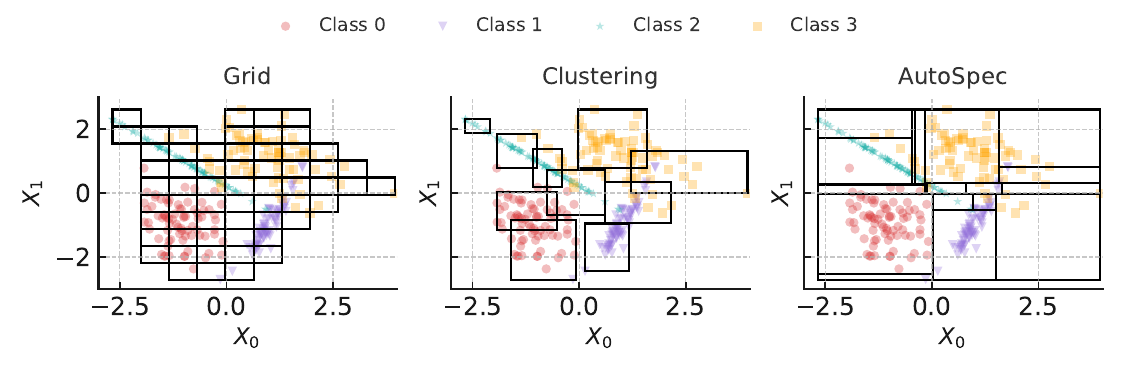}
  \caption{Visualization of the specification set generated by different algorithms.}
  \label{fig:spec-gen-algo-visualization}
  \vspace{-10pt}
\end{figure}

\heading{Illustrations.} We present a visualization of the specifications generated by different algorithms on a 2D dataset in Figure~\ref{fig:spec-gen-algo-visualization} (specifications as rectangles, data points as dots, and different colors for different points’ corresponding
classes). This illustration highlights that \sys achieves the most comprehensive, accurate, and tight coverage of the input space compared to other methods. The effectiveness and accuracy of these generated specifications are further evaluated in \autoref{s:eval}.

%% file: sections/fig-overview.tex
\begin{figure}[t]
\centering
\includegraphics[width=1.0\textwidth]{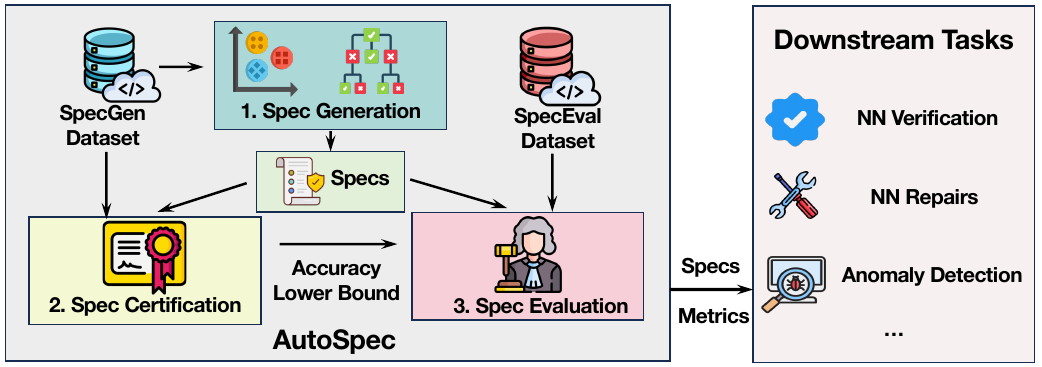}
\caption{
Overview of \sys's workflow. 
\sys first splits the labeled data into separate sets for specification generation and evaluation. 
The \emph{Specification Generation Framework} constructs candidate specifications from the generation dataset (\autoref{s:specgen}). 
The \emph{Specification Certification Framework} then certifies a formal accuracy bound for each specification based on empirical evidence (\autoref{s:pac-certify}). 
Finally, the \emph{Specification Evaluation Framework} computes interpretable quality metrics—such as comprehensiveness and accuracy—on the evaluation dataset (\autoref{s:speceval}).
The resulting certified and evaluated specifications can be directly used for downstream verification tasks.
}
\label{fig:overview}
\vspace{-0.1in}
\end{figure}

%% file: sections/03-3-specification-bound.tex
\mysubsection{Specification Accuracy Certification}
\label{s:pac-certify}

After generating the set of candidate specifications  
\(
\Phi = \{\phi_1, \dots, \phi_{|\Phi|}\},
\)
our next goal is to certify that the empirical accuracy measured for each specification reliably reflects its true accuracy on the underlying data distribution. We accomplish this by leveraging Hoeffding’s inequality to provide a probably approximately correct (PAC) lower bound for each individual specification.

For any specification $\phi = (\phi_X, \phi_Y)$ and sample $(x_i, y_i) \in \mathcal{D}_{\mathrm{gen}}$, define the Bernoulli random variable:
\[
X_i(\phi) = \mathbf{1}\left\{x_i \in \phi_X \wedge y_i \notin \phi_Y \right\}.
\]
The empirical error rate is $\hat{p}(\phi) = \frac{1}{N} \sum_{i=1}^{N} X_i(\phi)$, and the empirical accuracy is $\hat{\alpha}(\phi) = 1 - \hat{p}(\phi)$, where $N = |\mathcal{D}_{\mathrm{gen}}|$. The true error rate is $\mathbb{E}[X_i(\phi)]$, and the true accuracy is $\alpha(\phi) = 1 - \mathbb{E}[X_i(\phi)]$.

\begin{proposition}[PAC Accuracy Bound]
      \label{prop:accuracy}
      Let $\mathcal{D}_{\mathrm{gen}}$ be a dataset of $N$ i.i.d. samples and fix any specification $\phi\in\Phi$. For any $\delta \in (0,1)$, define 
      \[
      \varepsilon_N(\delta) = \sqrt{\frac{\ln(1/\delta)}{2N}}.
      \]
      Then, with probability at least $1-\delta$ (over the draw of the dataset),
      \[
      \alpha(\phi) \geq \hat{\alpha}(\phi) - \varepsilon_N(\delta),
      \]
      where $\hat{\alpha}(\phi)$ is the empirical accuracy and $\alpha(\phi)$ is the true accuracy of $\phi$.
      \end{proposition}
      
      \begin{proof}
      For any fixed $\phi$, the random variables $X_i(\phi) \in [0,1]$ are independent. By the one-sided Hoeffding inequality,
      \[
      \Pr\left( p(\phi) - \hat{p}(\phi) \geq \varepsilon_N(\delta) \right) \leq \delta,
      \]
      where $p(\phi) = \mathbb{E} X_i(\phi)$ and $\hat{p}(\phi) = \frac{1}{N} \sum_{i=1}^N X_i(\phi)$.  
      Rearranging gives, with probability at least $1-\delta$,
      \[
      p(\phi) \leq \hat{p}(\phi) + \varepsilon_N(\delta),
      \]
      so
      \[
      \alpha(\phi) = 1 - p(\phi) \geq 1 - \hat{p}(\phi) - \varepsilon_N(\delta) = \hat{\alpha}(\phi) - \varepsilon_N(\delta).
      \]
      \end{proof}

This guarantee depends only on the sample size $N$ and chosen confidence level $\delta$, not on the size or coverage of $\phi_X$. For example, with $N = 50{,}000$ and $\delta = 0.05$, $\varepsilon_N(\delta) \approx 0.547\%$, meaning that an empirical accuracy of $97\%$ certifies at least $96.45\%$ true accuracy with $95\%$ confidence.

This PAC certificate allows us to translate empirical results into rigorous guarantees for each generated specification, facilitating reliable downstream use and formal verification.

%% file: sections/03-1-evaluation-metrics.tex
\mysubsection{Specification Evaluation Framework}
\label{s:speceval}

In this part, we introduce the specification evaluation framework in \sys, carefully defining a set of standardized metrics, such as precision, recall, and the new Pass Rate, to quantitatively assess the quality of \specs. These metrics allow for effective comparisons of different specification generation algorithms (\autoref{s:specgen}).

Our metric design follows three principles:

\begin{myenumerate2}

\item \emph{Comprehensive coverage}: The aggregate input ranges of \specs should span a wide range of the input space. Broad coverage ensures effective regulation of model behavior across various scenarios, providing reliable guarantees for learning-augmented systems.

\item \emph{High accuracy}: Specifications should closely align with ground-truth data points, ensuring theoretical correctness and practical relevance to real-world scenarios.

\item \emph{Tight constraint}: Specifications should impose relatively narrow \textit{output ranges} to meaningfully restrict model behavior. Overly permissive output ranges undermine their utility in verification.

\end{myenumerate2}

\heading{Specification Evaluation Inputs and Outputs.}
Specification evaluation takes two inputs: (1) a \spec set $\Phi$, consisting of specifications $\phi_i = (\phi_{X_i}, \phi_{Y_i})$, and (2) an evaluation dataset comprising data points $X_{eval}$ and corresponding ground-truth labels $Y_{eval}$. The evaluation outputs metrics quantifying the overall quality and effectiveness of $\Phi$, providing insights into the specifications' coverage and accuracy.

\heading{Specification Metrics.} We define True Positives (TP), False Positives (FP), and False Negatives (FN) to evaluate specifications:

\begin{myitemize2}
\item \emph{True Positive (TP)}: A data point $(x_i,y_i)$ satisfies a specification $\phi_j\in\Phi$ if $x_i \in \phi_{X_j} \wedge y_i \in \phi_{Y_j}$.
\item \emph{False Positive (FP)}: A data point $(x_i,y_i)$ is covered by a specification $\phi_j\in\Phi$ but does not meet its output constraints, i.e., $x_i \in \phi_{X_j} \wedge y_i \notin \phi_{Y_j}$.
\item \emph{False Negative (FN)}: A data point $(x_i,y_i)$ not covered by any specification in $\Phi$, i.e., $x_i \notin \phi_{X_j}$ for all $\phi_j \in \Phi$.
\end{myitemize2}

True negatives (TN) are omitted as they carry limited practical meaning here.

\heading{Precision, Recall, and F1 Score.} Based on these metrics, we calculate:

\begin{myitemize2}
\item \textit{Precision} = TP / (TP + FP), measuring overall specification accuracy.
\item \textit{Recall} = TP / (TP + FN), quantifying coverage of the input space.
\item \textit{F1 score} is the harmonic mean of precision and recall, offering a single consolidated metric.
\end{myitemize2}

\heading{Pass Rate.} Utilizing the PAC bound from \autoref{s:pac-certify}, we formally define the Pass Rate as follows. For each specification $\phi$, we randomly draw $M$ subsets from the evaluation set and compute the empirical accuracy $\tilde{\alpha}(\phi)$ for each subset. Let $M'$ denote the number of subsets (each of size $S$) where $\tilde{\alpha}(\phi) \geq \hat{\alpha}(\phi) - \varepsilon_N(\delta)$. A specification passes if $\frac{M'}{M} \geq 1 - \delta$. The Pass Rate is then defined as:

$$
\textit{PassRate}(\Phi) = \frac{1}{|\Phi|}\sum_{\phi \in \Phi}\mathbf{1}\left\{\frac{M'}{M} \geq 1 - \delta\right\}.
$$

The Pass Rate thus reflects the fraction of specifications that consistently meet or exceed their certified statistical accuracy guarantees under repeated evaluation. By reporting the Pass Rate, we can empirically verify whether the PAC bounds derived during training remain valid when tested on new data. A Pass Rate close to $1$ indicates that the statistical guarantees are both meaningful and robust—most specifications perform at least as well as promised in practice. Conversely, if the Pass Rate is noticeably lower than $1$, it may suggest overfitting, insufficient data, or distribution shift between the training and evaluation sets.

In our evaluation, we utilize the Pass Rate to: (1) validate the calibration of our certified guarantees, (2) compare the reliability of different specification generation algorithms, and (3) identify scenarios where the guarantees may not generalize. This makes Pass Rate a practical and interpretable metric for assessing both the statistical soundness and real-world reliability of specification sets.

\heading{Addressing Corner Cases.} Corner cases, such as overlapping specifications and excessively large output ranges, require special handling:

\emph{Overlapping specifications.} A data point covered by multiple overlapping specifications is TP only if its output satisfies all overlapping $\phi_{Y_j}$; otherwise, it is FP.

\emph{Unbounded output ranges.} Specifications with excessively large output ranges are not helpful for verification due to being overly permissive. In the extreme case, if a specification defines an infinite output range, it would be trivially satisfied by any neural network, rendering verification useless. To rectify this, we discard specifications with overly broad or infinite output ranges prior to evaluation. For classification, output ranges must specify exactly one class. For regression, we require $\phi_{Y_j}^{\max} - \phi_{Y_j}^{\min} \le \alpha (Y_{\max}-Y_{\min})$, with $\alpha \in (0,1)$, to ensure meaningful constraints.

%% file: sections/05-evaluation.tex
\mysection{Experimental Evaluation}
\label{s:eval}

In this section, we conduct extensive evaluations on four diverse datasets to address the following questions:
\begin{myitemize2}
    \item How well does \sys perform across different datasets? Are the certified accuracy lower bounds reliable, and do the evaluation metrics effectively distinguish between various \spec generation algorithms?
    \item How do automatically generated \specs compare to human-designed specifications?
    \item Are auto-generated \specs useful for downstream tasks?
\end{myitemize2}

\begin{table*}[t]
\centering
\resizebox{\textwidth}{!}{%
\begin{tabular}{cccccccccc}
\toprule
\multirow{2.5}{*}{\textbf{Application}} & \multirow{2.5}{*}{\textbf{Input Dimension}} & \multirow{2.5}{*}{\textbf{Methods}} & \multicolumn{6}{c}{\textbf{Metrics}} \\ \cmidrule{4-10} 
 &  &  & \textit{\#TP} & \textit{\#FP} & {\textit{\#FN}} & \textit{Precision (\%)} & {\textit{Recall (\%)}} & \textit{F1 (\%)} & \textit{Pass Rate (\%)} \\ 
\midrule
\multirow{3}{*}{Toy Spiral Data} & \multirow{3}{*}{2} & Grid & 507 & 27 & 6 & \underline{94.94} & \underline{98.83} & \underline{96.85} & \underline{99.79} \\
 &  & KMeans & 445 & 29 & 66 & 93.88 & 87.08 & 90.36 & 99.47 \\
 &  & \cellcolor{teal!10} \dt & \cellcolor{teal!10} 536 & \cellcolor{teal!10} 4 & \cellcolor{teal!10} 0 & \cellcolor{teal!10} \textbf{99.26} & \cellcolor{teal!10} \textbf{100.00} & \cellcolor{teal!10} \textbf{99.63} & \cellcolor{teal!10} \textbf{100.00} \\
\midrule
\multirow{4}{*}{Throughput Prediction} & \multirow{4}{*}{4} 
& \cellcolor{pink} Human & \cellcolor{pink} 294{,}505 & \cellcolor{pink} 34{,}121 & \cellcolor{pink} 463{,}278 & \cellcolor{pink} \textbf{89.62} & \cellcolor{pink} 38.86 & \cellcolor{pink} 54.22 & \cellcolor{pink} \underline{99.96} \\
 & & Grid & 612{,}015 & 176{,}663 & 3{,}226 & \underline{77.60} & 99.48 & \textbf{87.19} & 99.5 \\
 &  & KMeans & 550{,}910 & 240{,}141 & 853 & 69.64 & \underline{99.85} & 82.05 & 97.15 \\
 &  & \cellcolor{teal!10} \dt & \cellcolor{teal!10} 569{,}134 & \cellcolor{teal!10} 222{,}587 & \cellcolor{teal!10} 183 & \cellcolor{teal!10} 71.89 & \cellcolor{teal!10} \textbf{99.97} & \cellcolor{teal!10} \underline{83.63} & \cellcolor{teal!10} \textbf{100.00} \\
\midrule
\multirow{3}{*}{Intrusion Detection} & \multirow{3}{*}{78} & \grid & \NaN & \NaN & {\NaN} & \NaN & {\NaN} & \NaN & \NaN\\
 &  & \kmeans & 703{,}663 & 87{,}124 & 7{,}327 & \underline{88.98} & \underline{98.97} & \underline{93.71} & \underline{99.96} \\ %
 &  & \cellcolor{teal!10} \dt & \cellcolor{teal!10} 797{,}826 & \cellcolor{teal!10} 288 & \cellcolor{teal!10} 0 & \cellcolor{teal!10} \textbf{99.96} & \cellcolor{teal!10} \textbf{100.00} & \cellcolor{teal!10} \textbf{99.98} & \cellcolor{teal!10} \textbf{100.00} \\
\midrule
\multirow{3}{*}{Beam Management} & \multirow{3}{*}{4096} & \grid & \NaN & \NaN & {\NaN} & \NaN & {\NaN} & \NaN & \NaN\\
 &  & \kmeans & 4{,}087 & 6{,}652 & 34{,}261 & \underline{38.05} & \underline{10.65} & \underline{16.65} & \underline{99.62} \\ %
 &  & \cellcolor{teal!10} \dt & \cellcolor{teal!10} 19{,}450 & \cellcolor{teal!10} 25{,}550 & \cellcolor{teal!10} 0 & \cellcolor{teal!10} \textbf{43.22} & \cellcolor{teal!10} \textbf{100.00} & \cellcolor{teal!10} \textbf{60.36} & \cellcolor{teal!10} \textbf{100.00} \\ 
\bottomrule
\end{tabular}%
}
\caption{Evaluation results across datasets. \textbf{Bold} and \underline{underline} indicate the best and second-best results for each dataset.}
\label{tab:overall-evaluation}
\vspace{-0.1in}
\end{table*}

\input{sections/05-1-experimental-setup}

\mysubsection{\sys's performance}

\begin{figure*}[t]
\centering
\minipage{0.48\textwidth}
  \centering
  \includegraphics[width=\linewidth]{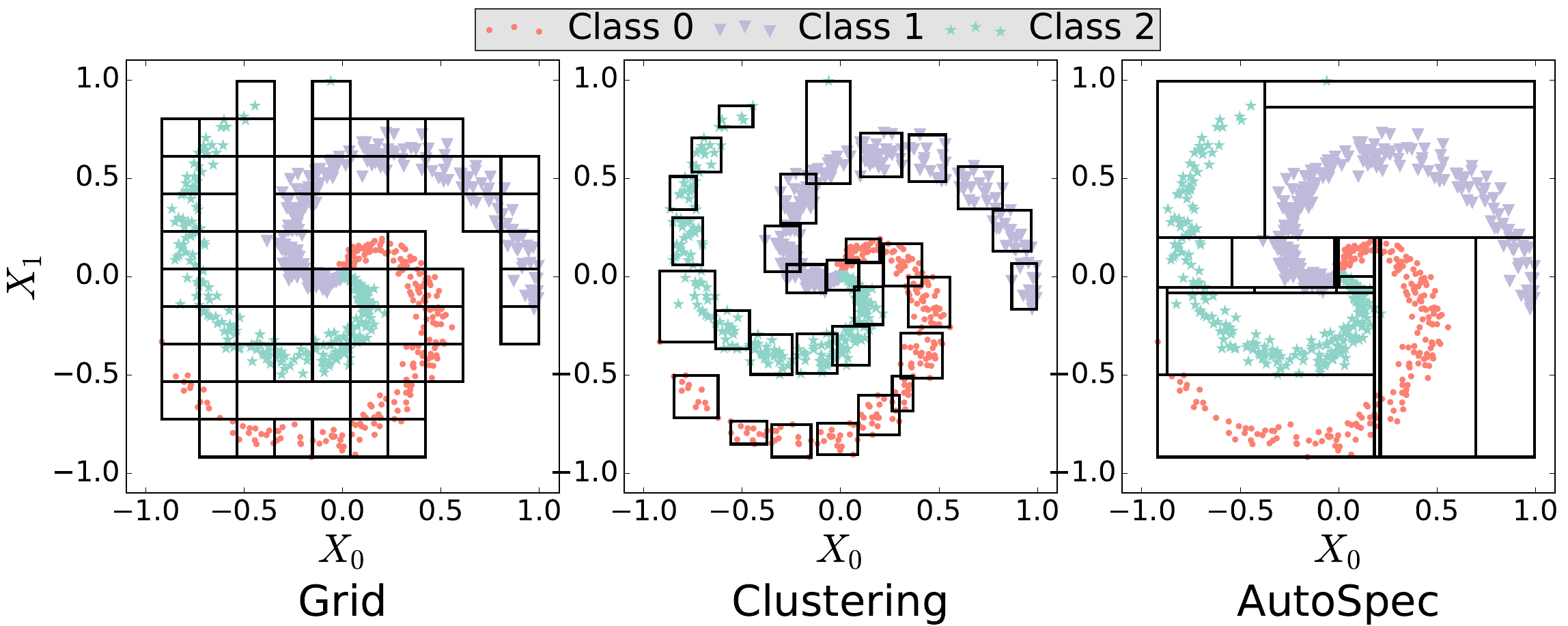}
  \vspace{-10pt}
  \caption{Specification visualization on the specification generation dataset.}
  \label{fig:visualization-spec-gen}
\endminipage\hfill
\minipage{0.48\textwidth}
  \centering
  \includegraphics[width=\linewidth]{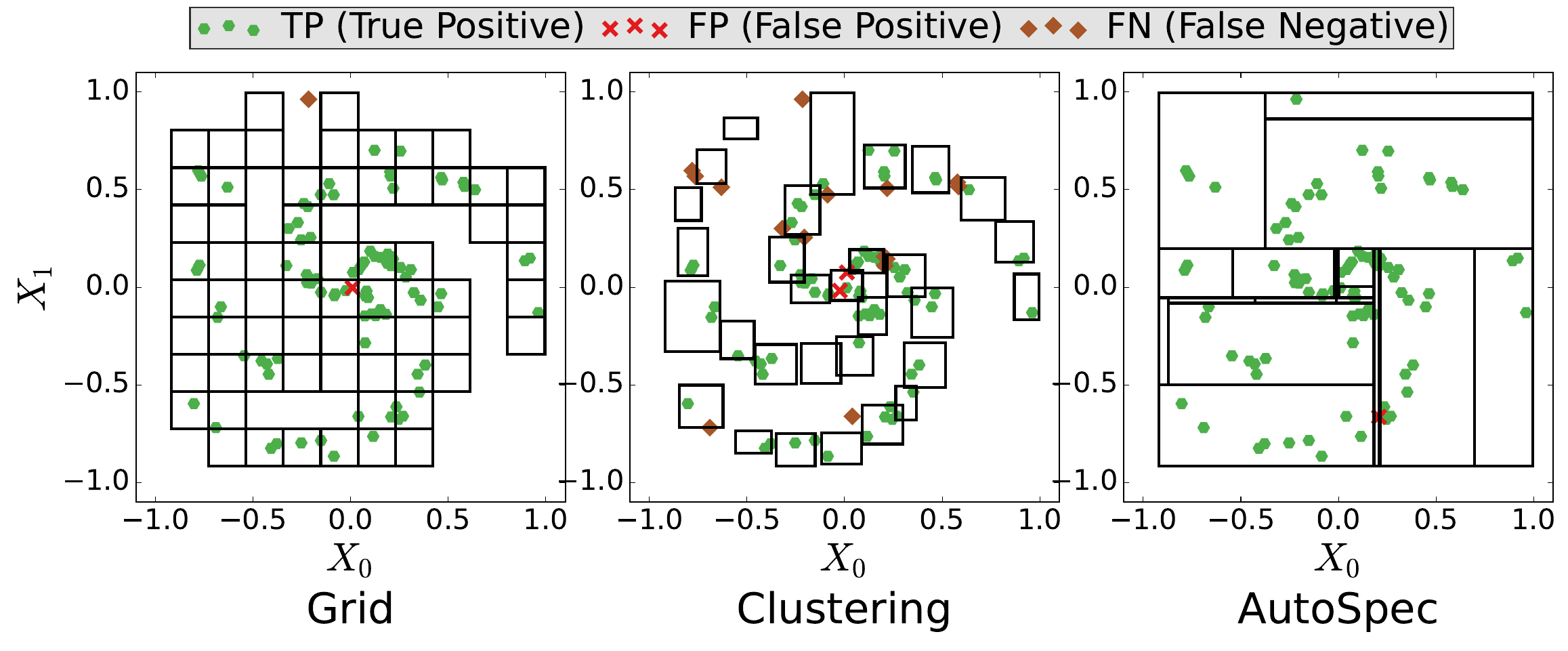}
  \vspace{-13pt}
  \caption{Specification visualization on the specification evaluation dataset.}
  \label{fig:visualization-spec-eval}
\endminipage\hfill
\vspace{-0.1in}
\end{figure*}

\input{sections/05-2-evaluation-results}

\mysubsection{Comparison with human-defined specifications}

We compare \sys-generated specifications against human-defined
ones for the throughput prediction dataset. To accommodate the tight output constraint in specifications, continuous
throughput values are divided into ten intervals (0--9), with each interval
representing a percentage range of the maximum throughput (e.g., 0 for 0--10\%
of Throughput$_{\text{max}}$). This allows us to only have to choose a label to
represent the output range for a specification.

Human-defined specifications are constructed using two principles:
(a) \emph{monotonic trend}—when historical throughputs increase or decrease monotonically, the current throughput is expected to follow the same trend, estimated via linear regression;
(b) \emph{stable trend}—if historical throughput remains stable, the next value is likely similar.

\begin{figure}[t]
\centering
\begin{minipage}{0.48\columnwidth}
\centering
\resizebox{\columnwidth}{!}{%
\begin{tabular}{cccccc}
\toprule
\textbf{Type} & \textbf{x{[}0{]}} & \textbf{x{[}1{]}} & \textbf{x{[}2{]}} & \textbf{x{[}3{]}} & \textbf{y (Prediction)} \\ \midrule
Monotonic Increase & 0 & 2 & 4 & 6 & 8 \\
Monotonic Decrease & 9 & 7 & 5 & 3 & 1 \\
Stable             & 0 & 0 & 0 & 0 & 0 \\
\bottomrule
\end{tabular}%
}
\caption{Examples of human-defined specifications.}
\label{tab:human-specification}
\end{minipage}
\hfill
\begin{minipage}{0.48\columnwidth}
\centering
\includegraphics[width=\linewidth]{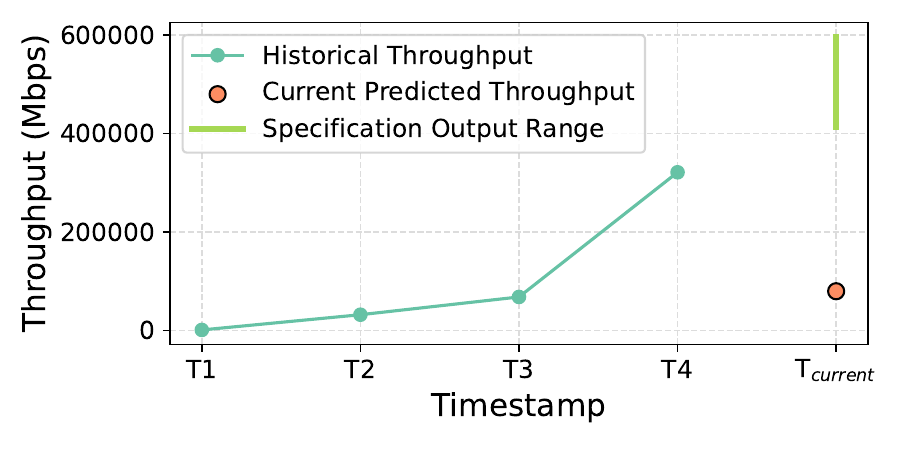}
\caption{Monotonic increase counterexample.}
\label{fig:increase-counterexample}
\end{minipage}
\vspace{-0.2in}
\end{figure}

The human-defined \spec set is constructed by selecting combinations that meet the monotonic and stable trend criteria (see Appendix~\ref{sec:appendix-human-defined-specification}, Table~\ref{tab:human-specification}). As shown in \autoref{tab:overall-evaluation}, \sys achieves a 53\% higher F1 score than human-defined \specs. While human \specs have high precision, they suffer from low recall, reflecting the limited coverage achievable by manual design.

\mysubsection{Verifying DNNs with generated \specs}

\textbf{Verification Engine.} To verify neural networks with the generated specifications,
we use \ttt{auto\_LiRPA}~\cite{xu2020automatic, xu2021fast,
wang2021beta} as our verification engine, which allows automatic bound
derivation and computation for general computational graphs.

\heading{Experiment Setup.}
The goal is to verify a trained neural network
against \sys-generated \specs
to identify potential vulnerabilities.
For the experiment, we use the same dataset---throughput prediction
dataset---for both training the model
and generating the specifications.
The model is a simple four-layer fully
connected neural network with ReLU, 
with 4, 10, 5, and 1 neurons in each layer. We terminate the training when its loss stabilizes and
loss variation \(\le 1 \times 10^{-3}\).
Then, we start the verification.
If the model fails the verification,
we extract data points that violate the \specs.
These points, termed as counterexamples, highlight
cases when the model's behavior deviates from the expected outcomes.

\heading{Detected Anomalies.}
An illustrative counterexample is presented in
Figure~\ref{fig:increase-counterexample}.
Analysis of the historical throughput
values suggests a monotonically increasing trend. However, the model's
predictions, contrastingly, indicate a substantial decrease, as depicted by the
red point (the specification's expected output range is described by
the green line). This discrepancy not only flags an anomaly but also exposes
underlying vulnerabilities within the model. We discuss additional
counterexamples in Appendix~\ref{sec:appendix-counterexample}.

%% file: sections/05-1-experimental-setup.tex
\mysubsection{Experimental Setups}

\heading{Datasets.}
We evaluate our framework on four datasets: the Toy Spiral dataset (a synthetic 2D classification benchmark), the Uplink Throughput Prediction dataset from the Colosseum O-RAN Dataset~\cite{bonati2021intelligence}, the CIC-IDS2017 Intrusion Detection dataset~\cite{sharafaldin2018toward}, and DeepBeam~\cite{polese2021deepbeam}, a 5G beam management dataset. Detailed descriptions of each dataset can be found in \autoref{s:appendix-dataset}.

Each dataset is split into three folds using a 6:1:3 ratio: the first portion is used for specification generation, the second for validation and parameter tuning, and the remaining portion for evaluating the generated specifications.

\heading{Metrics.}
For evaluation, we use the metrics defined in our evaluation framework (\autoref{s:speceval}): true positives (TP), false positives (FP), false negatives (FN), precision, recall, F1 score, and pass rate.

\heading{Baselines and Parameters.}
We compare our framework against two baseline methods: a fixed-size grid approach (``Grid'') and a clustering-based approach (``Clustering''). For the grid method, we set the number of bins $\beta$ to 10. For the clustering method, we use $k$-means, selecting the number of clusters as 30, 100, 1000, and 3000 for the spiral, uplink throughput, intrusion detection, and beam management tasks, respectively, based on validation results. For the decision tree method, we use maximum depths of 10, 30, 40, and 70 for the same tasks respectively.
For the throughput prediction task, we also include human-defined specifications; however, due to the high dimensionality of the input features, human experts were unable to define specifications for the intrusion detection and beam management tasks. To ensure that generated specifications maintain tight output constraints, we set the maximum allowable output range parameter $\alpha$ to $20\%$. In the certification procedure, the number of samples $M$ is set to 100, and the subset size $S$ is chosen as one-tenth of the test set size.

%% file: sections/05-2-evaluation-results.tex
\heading{Results.}
We run \sys and baselines on the four datasets.
Table \ref{tab:overall-evaluation} shows the results.
We observe that \sys performs the best on average
across all datasets and applications.
Overall,
\sys improves the F1 score over the second best by 68\% on average.

In the first two datasets,
where input dimensions are small,
the Grid algorithm also achieves good performance.
However, for intrusion detection and beam management,
due to the high-dimensional input space,
the \grid's time complexity grows exponentially,
and it times out when producing \specs. Specifically, the algorithm was allowed to run for 6 hours on a %
64-Core Processor before being terminated, yet it failed to produce \specs within this timeframe.
\kmeans algorithm performs worse than the \grid and \sys,
but it runs fast (in polynomial time) and can generate specifications
for high-dimensional spaces.
We also observe on the beam management dataset, that both \kmeans and
\sys's performance drops significantly.
This is because the problem is fundamentally hard with
an increased number of input dimensions.
This suggests future work of building algorithms
for high-dimensional inputs.

\heading{Illustration of Auto-Generated Specifications.}
To better understand the generated specifications, we visualize the toy spiral dataset in Figures~\ref{fig:visualization-spec-gen} and~\ref{fig:visualization-spec-eval}, showing specifications (rectangles), data points (dots), and their classes (colors). Figure~\ref{fig:visualization-spec-gen} shows that both \kmeans and \grid ignore label information when splitting the feature space, often producing specifications with mixed-class data and resulting in false positives during evaluation. In contrast, \sys partitions the input space more effectively, generating class-pure specifications with high recall and precision.

\heading{Effectiveness of PAC Accuracy Bound}
As shown by the consistently high Pass Rates in \autoref{tab:overall-evaluation}, our PAC accuracy bound reliably translates to real-world performance. AutoSpec achieves a perfect Pass Rate of $100\%$ across all datasets, confirming that nearly all generated specifications meet their certified accuracy guarantees on new data. Even baseline and human-defined specifications maintain high Pass Rates (higher than 97\%), demonstrating that the PAC accuracy bound provides robust and practical statistical guarantees for specification correctness.

%% file: sections/07-conclusion.tex
\mysection{Conclusion}

Formal verification of neural networks is essential for the safety and reliability of learning-augmented systems, but current methods are limited by the need for human-crafted specifications. This work introduces \sys, the first end-to-end framework for automated specification generation, certification, and evaluation. By leveraging data-driven methods and rigorous evaluation metrics, \sys produces reliable specifications with minimal human effort. Experiments across four domains show that \sys outperforms expert-defined and baseline approaches, establishing a new standard for scalable neural network verification.

We include discussions of limitations and broader impact in the \autoref{s:limitation} and \autoref{s:impact-statements}.

%% file: sections/08-appendix.tex
\input{sections/08-00-algorithms}
\input{sections/08-01-human-defined-specification}

\input{sections/08-02-counterexamples}

\input{sections/10-discussion}
\input{sections/09-impact-statement.tex}

%% file: sections/08-00-algorithms.tex
\mysection{Pseudo Code for Secification Extraction Primitive.} 
All generation algorithms invoke the same helper,
\textsc{ExtractSpec}, which converts a candidate rectangle
\(R\subseteq\mathbb{R}^{n}\) into a specification:

Given a candidate rectangle \(R\subseteq\mathbb{R}^{n}\),
\autoref{algo:spec-extract}

\begin{enumerate}[leftmargin=12pt,itemsep=2pt]
\item collects the subset
      \(\mathcal{D}_{R}=\{(x,y)\in\mathcal{D}_{\mathrm{gen}}\mid x\in R\}\);
\item for classification, returns the majority label;  
      for regression, returns one standard deviation about the mean;
\item discards \(R\) if \(\mathcal{D}_{R}=\varnothing\),
      guaranteeing every retained specification is data‑grounded.
\end{enumerate}

\vspace{-4pt}
\begin{algorithm}[h]
\caption{\textsc{ExtractSpec}$(R,\mathcal{D}_{\mathrm{gen}},\text{task})$}
\label{algo:spec-extract}
$\mathcal{D}_{R}\leftarrow\{(x,y)\in\mathcal{D}_{\mathrm{gen}}\mid x\in R\}$\;
\If{$\mathcal{D}_{R}=\varnothing$}{\Return \textsc{None}}
\uElseIf{$\text{task}=\textsc{cls}$}{
  $y^{\star}\leftarrow\arg\max_{c}\#\{y=c\mid(x,y)\in\mathcal{D}_{R}\}$\;
  \Return $(R,\{y^{\star}\})$\;
}
\Else(\textsc{reg}){
  $\mu,\sigma\leftarrow\text{mean,std of }\{y\mid(x,y)\in\mathcal{D}_{R}\}$\;
  \Return $(R,[\mu-\sigma,\;\mu+\sigma])$\;
}
\end{algorithm}

\mysection{Pseudo code for specification generation algorithms}
\label{s:pseudo-code}

In this section, we present the pseudo-code for grid-based and clustering-based specification generation algorithms.

\mysubsection{Grid-Based}
\label{ss:grid}

The grid-based method is a straightforward approach to specification generation.
It divides the input space into fixed-size hyperrectangular cells by splitting each feature axis into $\beta$ equal bins, forming \(\beta^{n}\) hyper‑rectangular cells.

\textbf{Algorithm Steps.} The execution steps of grid-based method is as follows. The complete implementation is detailed in Algorithm \ref{algo:grid-with-extract}.
\begin{enumerate}[leftmargin=12pt,itemsep=2pt]
\item For every feature \(p=1,\dots,n\) compute
      \(l_{p}^{\min}=\min_{i}x_{i,p}\) and
      \(u_{p}^{\max}=\max_{i}x_{i,p}\).
\item Set step size
      \(t_{p}=(u_{p}^{\max}-l_{p}^{\min})/\beta\).
\item Enumerate all cells
      \(R=\prod_{p=1}^{n}[\,l_{p}^{\min}+k_{p}t_{p},\;
                           l_{p}^{\min}+(k_{p}+1)t_{p})\)
      with \(k_{p}\in\{0,\dots,\beta-1\}\);
      apply \textsc{ExtractSpec} to each non‑empty \(R\).
\end{enumerate}

\textbf{Complexity.} The algorithm runs in $O(\beta^n + N)$ time, where $N$ is the dataset size. This becomes exponential in the input dimension $n$.

\textbf{Limitations.} While conceptually simple, this approach has several drawbacks. The fixed bin size cannot adapt to varying data densities across the input space. Cells may contain mixed labels, resulting in imprecise output constraints. Many cells remain empty, wasting computational resources. Furthermore, the exponential complexity makes it impractical for high-dimensional data.

\begin{algorithm}[h]
\caption{\textsc{Grid-based Generation}}
\label{algo:grid-with-extract}
\KwIn{$\mathcal{D}_{\mathrm{gen}}=\{(x_i,y_i)\}_{i=1}^{N}$,\; bins $\beta$,\; task (\textsc{cls} or \textsc{reg})}
\KwOut{Specification set $\Phi$}
\tcp{Compute per‑feature min/max and step size}
\For{$p\gets 1$ \KwTo $n$}{
  $l_p^{\min}\gets\min_i x_{i,p}$;\;
  $u_p^{\max}\gets\max_i x_{i,p}$;\;
  $t_p\gets(u_p^{\max}-l_p^{\min})/\beta$\;
}
$\Phi\leftarrow\emptyset$\;
\tcp{Enumerate every cell in the $\beta^n$ grid}
\For(\tcp*[f]{multi‑index over bins}){$k_1,\dots,k_n\in\{0,\dots,\beta-1\}$}{
  $R\gets\prod_{p=1}^{n}\bigl[l_p^{\min}+k_pt_p,\;l_p^{\min}+(k_p+1)t_p\bigr)$\;
  $\phi\gets\textsc{ExtractSpec}(R,\mathcal{D}_{\mathrm{gen}},\text{task})$\;
  \If{$\phi\neq\textsc{None}$}{ $\Phi\leftarrow\Phi\cup\{\phi\}$ }
}
\Return $\Phi$
\end{algorithm}

\mysubsection{Density‑Aware Clustering}
\label{ss:cluster}

The clustering-based method aims to address the limitations inherent in the fixed-size grid approach, especially its rigidity in adapting to diverse data distributions. 

\textbf{Algorithm Steps.} The execution steps of clustering-based method is as follows. The complete implementation is detailed in \autoref{algo:cluster}.
\begin{enumerate}[leftmargin=12pt,itemsep=2pt]
      \item Apply clustering algorithms (e.g., $k$-means or DBSCAN) to identify natural clusters \(C_1, C_2, \ldots, C_k\) based on data similarity
      \item For each cluster \(C_i\), compute the hyperrectangle \(R_{C_i}\) by finding the minimum \(x^{\text{min}}_{C_i}\) and maximum \(x^{\text{max}}_{C_i}\) boundaries across each dimension
      \item Mathematically, \(R_{C_i} = \prod_{j=1}^{n} [x^{\text{min}}_{C_i,j}, x^{\text{max}}_{C_i,j}]\), where \(n\) is the number of features
      \item Apply \textsc{ExtractSpec} to each cluster's hyperrectangle to generate specifications
\end{enumerate}

\textbf{Complexity.} For Lloyd-style $k$-means, the algorithm runs in $O(kNd)$ time, where $k$ is the number of clusters ($k \ll N$), $N$ is the dataset size, and $d$ is the number of dimensions. This is significantly more scalable than the grid method's exponential complexity.

\textbf{Limitations.} While more adaptive than the grid approach, the clustering method still has drawbacks. Since clustering only considers input features and ignores labels, a single cluster may contain data points from multiple classes or with high variance in the label space. This can lead to imprecise output constraints.

\begin{algorithm}[h]
  \caption{\textsc{ClusterSpecGeneration}}
  \label{algo:cluster}
  \KwIn{$\mathcal{D}_{\mathrm{gen}}$, clustering routine (\textit{e.g.}, $k$‑means), task}
  \KwOut{Specification set $\Phi$}
  Run clustering on $\{x_i\}$ to obtain clusters $C_1,\dots,C_k$\;
  $\Phi\leftarrow\emptyset$\;
  \ForEach{$C_j$}{
    $R_{C_j}\gets\prod_{p=1}^{n}
       \bigl[\min_{x\in C_j}x_p,\;\max_{x\in C_j}x_p\bigr]$\;
    $\phi\gets\textsc{ExtractSpec}(R_{C_j},\mathcal{D}_{\mathrm{gen}},\text{task})$\;
    \If{$\phi\neq\textsc{None}$}{ $\Phi\leftarrow\Phi\cup\{\phi\}$ }
  }
  \Return $\Phi$
  \end{algorithm}

\mysubsection{\sys: Decision Tree-based Generation}
\sys core algorithm is the decision-tree based generation algorithm. It train a CART tree on $\mathcal{D}_{\mathrm{gen}}$ using Gini impurity
(classification) or squared‑error reduction (regression).  
Each internal node stores a predicate $x_{p}\!\le\!t$, thus slicing the
input space into finer rectangles.  Every leaf defines a rectangle
$R_{L}$ by intersecting predicates along its path; we convert each
$R_{L}$ into a specification via \textsc{ExtractSpec}.

\begin{algorithm}[h]
  \caption{\sys: Decision‑Tree–Based Specification Generation}
  \label{algo:autospec}
  \DontPrintSemicolon
  \KwIn{Generation set $\mathcal{D}_{\mathrm{gen}}\!=\!\{(x_i,y_i)\}_{i=1}^{N}$,\;
        task $\in\{\textsc{cls},\textsc{reg}\}$}
  \KwOut{Specification set $\Phi$}
  
  \BlankLine
  \textbf{Train tree}\;
  \Indp Train a CART decision tree $T$ on $\mathcal{D}_{\mathrm{gen}}$\;
  \Indm
  \BlankLine
  $\Phi \leftarrow \varnothing$\tcp*{accumulator for specs}
  
  \ForEach(\tcp*[h]{DFS}){leaf $L$ in $T$}{
      \textbf{Accumulate path predicates}\;
      \Indp
      Initialise lower bounds $\ell[1\!:\!n]\leftarrow -\infty$, upper bounds $u[1\!:\!n]\leftarrow +\infty$\;
      \ForEach{predicate $C$ on the root$\!\rightarrow\!L$ path}{
          Parse $C$ as $(f,\;\bowtie,\;t)$ where $f$ is a feature index and $t$ a threshold\;
          \uIf{$\bowtie$ is $\le$ or $<$}{
              $u[f] \leftarrow \min\{u[f],\,t\}$\;
          }\Else(\tcp*[f]{$>$ or $\ge$}){
              $\ell[f] \leftarrow \max\{\ell[f],\,t\}$\;
          }
      }
      \Indm
      \textbf{Form rectangle}\;
      $R_{L} \leftarrow \displaystyle\prod_{j=1}^{n}[\ell[j],\,u[j]]$\;
      \textbf{Convert to specification}\;
      $\phi \leftarrow \textsc{ExtractSpec}(R_{L},\mathcal{D}_{\mathrm{gen}},\text{task})$\;
      \If{$\phi \neq \textsc{None}$}{ $\Phi \leftarrow \Phi \cup \{\phi\}$ }\;
  }
  \Return $\Phi$\;
  \end{algorithm}

  \mysection{Dataset Details}
  \label{s:appendix-dataset}
  In our experiments, we evaluate our framework on four datasets: one synthetic toy dataset for visualization and three real-world datasets from different system domains.
  
  \begin{itemize}
      \item \emph{Toy Spiral Dataset.} The spiral dataset~\cite{spiraldataset} is a synthetic 2D classification benchmark designed for visualization. It contains three classes, each with 300 data points, where the task is to predict the class given a 2D location $(X_0, X_1)$. The resulting specifications have two-dimensional input constraints and a single-dimensional output constraint.
  
      \item \emph{Uplink Throughput Prediction Dataset.} Throughput prediction is critical for network resource management~\cite{kumar2018workload, chien2019dynamic, fu2022traffic}. We use the Colosseum O-RAN dataset~\cite{bonati2021intelligence}, which provides base station uplink throughput traces under various conditions. The target task is time series forecasting: the model uses four historical values to predict the current throughput. Specifications for this regression problem have four-dimensional input constraints and a one-dimensional output constraint.
  
      \item \emph{Intrusion Detection Dataset.} Intrusion detection and prevention systems (IDSs/IPSs) are essential for protecting cyber environments. We use the CIC-IDS2017 dataset~\cite{sharafaldin2018toward}, which includes labeled network flows for both benign and multiple attack types. Each flow is represented by a 78-dimensional feature vector, and the classification task is to identify the flow as one of nine possible labels (e.g., benign, brute force FTP, brute force SSH). Specifications are defined with 78-dimensional input constraints and a single output dimension.
  
      \item \emph{Beam Management Dataset.} Millimeter wave (mmWave) radios require precise beam management to maintain reliable connections. DeepBeam~\cite{polese2021deepbeam} introduces deep learning models for this purpose, and provides a dataset of signal samples (I/Q data) under varying transmission conditions, including different beams and receiver gain levels. The input consists of 4096-dimensional feature vectors, with the task of classifying the transmit beam. Thus, specifications here have 4096-dimensional input constraints and a one-dimensional output constraint.
  \end{itemize}

%% file: sections/08-01-human-defined-specification.tex
\mysection{Human Defined Specification}
\label{sec:appendix-human-defined-specification}

This section outlines examples of human-designed specifications. Each specification has been translated into actual throughput speeds, based on the highest observed throughput of 919264 Mbps. The labels from ``0'' to ``9'' represent different ranges of throughput. For example, label ``0'' means the throughput is between 0 and 91926.4 Mbps. Each label after ``0'' covers an equally sized higher range of speeds, all the way up to the top speed. We list parts of our specifications in Table~\ref{tab:human_defined_specs}.

\begin{table}[h!]
\centering
\begin{tabular}{cccccc}
\toprule
\textbf{Type} & \textbf{x[0]} & \textbf{x[1]} & \textbf{x[2]} & \textbf{x[3]} & \textbf{y (Prediction)} \\
\midrule
Stable & 0 & 0 & 0 & 0 & 0 \\
\midrule
Increasing & 0 & 1 & 2 & 3 & 4 \\
\midrule
Increasing & 0 & 1 & 2 & 4 & 5 \\
\midrule
Increasing & 0 & 1 & 2 & 5 & 6 \\
\midrule
Increasing & 0 & 1 & 2 & 6 & 7 \\
\midrule
Increasing & 2 & 4 & 5 & 7 & 8 \\
\midrule
Decreasing & 5 & 4 & 1 & 0 & 0 \\
\midrule
Decreasing & 5 & 4 & 2 & 0 & 0 \\
\midrule
Decreasing & 5 & 4 & 2 & 1 & 0 \\
\midrule
Decreasing & 9 & 8 & 7 & 6 & 5 \\
\midrule
Stable & 9 & 9 & 9 & 9 & 9 \\
\bottomrule
\end{tabular}
\vspace{4pt}
\caption{Examples of human-defined specifications.}
\label{tab:human_defined_specs}
\end{table}

%% file: sections/08-02-counterexamples.tex
\mysection{Counterexamples}
\label{sec:appendix-counterexample}

\begin{figure*}[t!]
\centering
\minipage{0.5\textwidth}
  \centering
  \includegraphics[width=\linewidth]{figs/increasing_11225.pdf}
  \caption{Monotonic increase counterexample.}
  \label{fig:counterexample-increasing}
\endminipage\hfill
\minipage{0.5\textwidth}
  \centering
  \includegraphics[width=\linewidth]{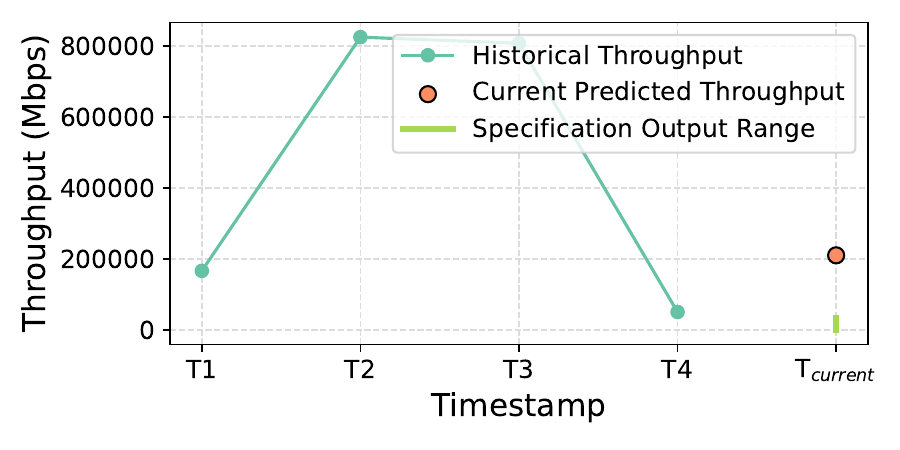}
  \caption{Monotonic decrease counterexample.}
  \label{fig:counterexample-decreasing}
\endminipage\hfill
\end{figure*}

We list two main counterexamples revealed by our generated specification set. Figure~\ref{fig:counterexample-increasing} illustrates an unexpected model behavior: although the anticipated trend for the current timestamp throughput is a continuous increase, the model's predicted output exhibits a substantial decrease. This deviation from the expected behavior is an anomaly. Conversely, Figure~\ref{fig:counterexample-decreasing} presents the reverse scenario. In this case, despite the projected trend indicating a decrease, the model's output shows an increase, further underscoring the discrepancies between the expected and actual model behaviors, revealing the vulnerabilities.

%% file: sections/10-discussion.tex
\section{Discussion}

In this section, we discuss some of the motivations and questions in this work.

\subsection{Why learn from data?}

In the context of machine learning systems, some previous works propose utilizing specifications to verify neural network safety before deployment into the system~\cite{eliyahu2021verifying, wei2023building}. However, all these previous works heavily rely on human experts to define the specification set, and these specification sets are often loose and can have errors. Considering the procedure of human experts in defining the specifications, they rely on their understanding of the system and their historical observation of the system behavior. Such an observation procedure can be considered as a step to learn from experience/data, which is a natural task for machine learning to automatically learn from data.

\subsection{Correctness of learned specifications}

Traditionally, specifications defined by domain experts have been considered as ground-truth laws that must not be violated. However, these expert-defined specifications are inherently limited by human experience and may occasionally be incorrect or incomplete. In contrast, our work proposes a data-driven approach to specification generation. While this method offers the potential to uncover patterns overlooked by human experts, it also introduces the risk of generating incorrect or spurious specifications, especially if the dataset contains noise or biases. To mitigate this risk, we could employ rigorous evaluation and filtering processes, selecting only those specifications that demonstrate a high probability of correctness based on our evaluation criteria. These high-confidence, data-derived specifications are then used to guide and constrain neural network behavior, potentially leading to more robust and reliable models. 

\subsection{Why Decision Trees?}

Our specification generation methods utilize decision trees due to their inherent advantages in this context. Decision trees naturally partition the input space, aligning with our goal of generating hyperrectangle based specifications. Their hierarchical structure offers interpretability, crucial for understanding and validating derived rules. Recent studies have shown decision trees can perform comparably to traditional neural networks in certain tasks \cite{meng2019pitree, zhuang2024learning}. 

%% file: sections/09-impact-statement.tex
\mysection{Limitations}
\label{s:limitation}
\subsection{Scalability to High-Dimensional Inputs}

While AutoSpec’s decision-tree–based approach is significantly more efficient than grid-based methods, its performance degrades with extremely high-dimensional input spaces. This is evident in the beam management task, where even AutoSpec shows reduced F1 scores due to the difficulty of constructing tight, accurate specifications in such large spaces (input dimensionality of 4096). Developing scalable generation methods tailored for high-dimensional settings remains an open challenge.
\subsection{Specification Expressiveness:}
AutoSpec generates specifications constrained to axis-aligned hyperrectangles in the input space and interval bounds in the output space. While this enables compatibility with state-of-the-art verifiers and interpretability, it also limits the expressiveness of the specifications. Complex input-output relationships may not be well captured by such simple geometric forms. Extending AutoSpec to support more expressive shapes (e.g., polytopes, zonotopes) could broaden its applicability.
\subsection{Dependence on Labeled Training Data:}
The quality of the generated specifications highly depends on the quality and representativeness of the labeled training data. If the training data distribution differs significantly from the real-world input distribution (i.e., distribution shift), the resulting specifications and their certified guarantees may no longer be valid or useful.

\mysection{Impact Statements}
\label{s:impact-statements}
The paper presents \sys, which holds substantial potential for enhancing the safety and reliability of machine learning models in safety-critical systems. By automating neural network specification generation, it mitigates the risks associated with manual specification processes, which are error-prone and often not comprehensive. This is particularly vital in domains where machine learning model behavior must be predictable and safe under diverse conditions. By providing a systematic approach to generating comprehensive and accurate specifications, this research contributes to the development of AI models that are more transparent and accountable.

%% file: reference.bib
@inproceedings{eliyahu2021verifying,
  title={Verifying learning-augmented systems},
  author={Eliyahu, Tomer and Kazak, Yafim and Katz, Guy and Schapira, Michael},
  booktitle={Proceedings of the 2021 ACM SIGCOMM 2021 Conference},
  pages={305--318},
  year={2021}
}

@online{spiraldataset,
    title={Spiral Toy Dataset},
    url={https://cs231n.github.io/neural-networks-case-study/},
    year={2022}
}

@article{bonati2021intelligence,
  title={Intelligence and learning in O-RAN for data-driven NextG cellular networks},
  author={Bonati, Leonardo and D'Oro, Salvatore and Polese, Michele and Basagni, Stefano and Melodia, Tommaso},
  journal={IEEE Communications Magazine},
  volume={59},
  number={10},
  pages={21--27},
  year={2021},
  publisher={IEEE}
}

@article{kumar2018workload,
  title={Workload prediction in cloud using artificial neural network and adaptive differential evolution},
  author={Kumar, Jitendra and Singh, Ashutosh Kumar},
  journal={Future Generation Computer Systems},
  volume={81},
  pages={41--52},
  year={2018},
  publisher={Elsevier}
}

@article{chien2019dynamic,
  title={Dynamic resource prediction and allocation in C-RAN with edge artificial intelligence},
  author={Chien, Wei-Che and Lai, Chin-Feng and Chao, Han-Chieh},
  journal={IEEE Transactions on Industrial Informatics},
  volume={15},
  number={7},
  pages={4306--4314},
  year={2019},
  publisher={IEEE}
}

@article{fu2022traffic,
  title={Traffic prediction-enabled energy-efficient dynamic computing resource allocation in cran based on deep learning},
  author={Fu, Yongqin and Wang, Xianbin},
  journal={IEEE Open Journal of the Communications Society},
  volume={3},
  pages={159--175},
  year={2022},
  publisher={IEEE}
}

@article{sharafaldin2018toward,
  title={Toward generating a new intrusion detection dataset and intrusion traffic characterization.},
  author={Sharafaldin, Iman and Lashkari, Arash Habibi and Ghorbani, Ali A},
  journal={ICISSp},
  volume={1},
  pages={108--116},
  year={2018}
}

@article{polese2021deepbeam,
  title={{DeepBeam: Deep Waveform Learning for Coordination-Free Beam Management in mmWave Networks}},
  author={Polese, Michele and Restuccia, Francesco and Melodia, Tommaso},
  journal={Proc. of ACM International Symposium on Mobile Ad Hoc Networking and Computing (MobiHoc)},
  year={2021}
}

@article{xu2020automatic,
  title={Automatic perturbation analysis for scalable certified robustness and beyond},
  author={Xu, Kaidi and Shi, Zhouxing and Zhang, Huan and Wang, Yihan and Chang, Kai-Wei and Huang, Minlie and Kailkhura, Bhavya and Lin, Xue and Hsieh, Cho-Jui},
  journal={Advances in Neural Information Processing Systems},
  volume={33},
  year={2020}
}

@inproceedings{xu2021fast,
    title={{Fast and Complete}: Enabling Complete Neural Network Verification with Rapid and Massively Parallel Incomplete Verifiers},
    author={Kaidi Xu and Huan Zhang and Shiqi Wang and Yihan Wang and Suman Jana and Xue Lin and Cho-Jui Hsieh},
    booktitle={International Conference on Learning Representations},
    year={2021},
    url={https://openreview.net/forum?id=nVZtXBI6LNn}
}

@article{wu2022scalable,
  title={Scalable verification of GNN-based job schedulers},
  author={Wu, Haoze and Barrett, Clark and Sharif, Mahmood and Narodytska, Nina and Singh, Gagandeep},
  journal={Proceedings of the ACM on Programming Languages},
  volume={6},
  number={OOPSLA2},
  pages={1036--1065},
  year={2022},
  publisher={ACM New York, NY, USA}
}

@article{wei2023building,
  title={Building Verified Neural Networks for Computer Systems with Ouroboros},
  author={Wei, Tianhao and Liu, Changliu and Jia, Zhihao and Tan, Cheng},
  journal={Proceedings of Machine Learning and Systems},
  volume={5},
  year={2023}
}

@inproceedings{bak2022neural,
  title={Neural network compression of ACAS Xu early prototype is unsafe: Closed-loop verification through quantized state backreachability},
  author={Bak, Stanley and Tran, Hoang-Dung},
  booktitle={NASA Formal Methods Symposium},
  pages={280--298},
  year={2022},
  organization={Springer}
}

@inproceedings{katz2019marabou,
  title={The marabou framework for verification and analysis of deep neural networks},
  author={Katz, Guy and Huang, Derek A and Ibeling, Duligur and Julian, Kyle and Lazarus, Christopher and Lim, Rachel and Shah, Parth and Thakoor, Shantanu and Wu, Haoze and Zelji{\'c}, Aleksandar and others},
  booktitle={Computer Aided Verification: 31st International Conference, CAV 2019, New York City, NY, USA, July 15-18, 2019, Proceedings, Part I 31},
  pages={443--452},
  year={2019},
  organization={Springer}
}

@article{wang2021beta,
  title={{Beta-CROWN}: Efficient bound propagation with per-neuron split constraints for complete and incomplete neural network verification},
  author={Wang, Shiqi and Zhang, Huan and Xu, Kaidi and Lin, Xue and Jana, Suman and Hsieh, Cho-Jui and Kolter, J Zico},
  journal={Advances in Neural Information Processing Systems},
  volume={34},
  year={2021}
}

@article{ammons2002mining,
  title={Mining specifications},
  author={Ammons, Glenn and Bodik, Rastislav and Larus, James R},
  journal={ACM Sigplan Notices},
  volume={37},
  number={1},
  pages={4--16},
  year={2002},
  publisher={ACM New York, NY, USA}
}

@inproceedings{lemieux2015general,
  title={General LTL specification mining (T)},
  author={Lemieux, Caroline and Park, Dennis and Beschastnikh, Ivan},
  booktitle={2015 30th IEEE/ACM International Conference on Automated Software Engineering (ASE)},
  pages={81--92},
  year={2015},
  organization={IEEE}
}

@inproceedings{tan2021building,
  title={Building verified neural networks with specifications for systems},
  author={Tan, Cheng and Zhu, Yibo and Guo, Chuanxiong},
  booktitle={Proceedings of the 12th ACM SIGOPS Asia-Pacific Workshop on Systems},
  pages={42--47},
  year={2021}
}

@inproceedings{kraska2018case,
  title={The case for learned index structures},
  author={Kraska, Tim and Beutel, Alex and Chi, Ed H and Dean, Jeffrey and Polyzotis, Neoklis},
  booktitle={Proceedings of the 2018 international conference on management of data},
  pages={489--504},
  year={2018}
}

@inproceedings{yan2020learning,
  title={Learning in situ: a randomized experiment in video streaming},
  author={Yan, Francis Y. and Ayers, Hudson and Zhu, Chenzhi and Fouladi, Sadjad and Hong, James and Zhang, Keyi and Levis, Philip and Winstein, Keith},
  booktitle={17th USENIX Symposium on Networked Systems Design and Implementation (NSDI 20)},
  pages={495--511},
  year={2020}
}

@article{liu2019machine,
  title={Machine learning and deep learning methods for intrusion detection systems: A survey},
  author={Liu, Hongyu and Lang, Bo},
  journal={applied sciences},
  volume={9},
  number={20},
  pages={4396},
  year={2019},
  publisher={mdpi}
}

@incollection{decima,
  title={Learning scheduling algorithms for data processing clusters},
  author={Mao, Hongzi and Schwarzkopf, Malte and Venkatakrishnan, Shaileshh Bojja and Meng, Zili and Alizadeh, Mohammad},
  booktitle={Proceedings of the ACM special interest group on data communication},
  pages={270--288},
  year={2019}
}

@inproceedings{geng2023towards,
  title={Towards reliable neural specifications},
  author={Geng, Chuqin and Le, Nham and Xu, Xiaojie and Wang, Zhaoyue and Gurfinkel, Arie and Si, Xujie},
  booktitle={International Conference on Machine Learning},
  pages={11196--11212},
  year={2023},
  organization={PMLR}
}

@inproceedings{le2018deep,
  title={Deep specification mining},
  author={Le, Tien-Duy B and Lo, David},
  booktitle={Proceedings of the 27th ACM SIGSOFT International Symposium on Software Testing and Analysis},
  pages={106--117},
  year={2018}
}

@article{zhuang2024learning,
  title={Learning a Decision Tree Algorithm with Transformers},
  author={Zhuang, Yufan and Liu, Liyuan and Singh, Chandan and Shang, Jingbo and Gao, Jianfeng},
  journal={arXiv preprint arXiv:2402.03774},
  year={2024}
}

@inproceedings{meng2019pitree,
  title={PiTree: Practical implementation of ABR algorithms using decision trees},
  author={Meng, Zili and Chen, Jing and Guo, Yaning and Sun, Chen and Hu, Hongxin and Xu, Mingwei},
  booktitle={Proceedings of the 27th ACM International Conference on Multimedia},
  pages={2431--2439},
  year={2019}
}

@article{chaudhary2024specification,
  title={Specification Generation for Neural Networks in Systems},
  author={Chaudhary, Isha and Lin, Shuyi and Tan, Cheng and Singh, Gagandeep},
  journal={arXiv preprint arXiv:2412.03028},
  year={2024}
}

@inproceedings{demarchi2023supporting,
  title={Supporting Standardization of Neural Networks Verification with VNNLIB and CoCoNet.},
  author={Demarchi, Stefano and Guidotti, Dario and Pulina, Luca and Tacchella, Armando and Narodytska, Nina and Amir, Guy and Katz, Guy and Isac, Omri},
  booktitle={FoMLAS@ CAV},
  pages={47--58},
  year={2023}
}
